\pdfoutput=1
\documentclass{article}

\newif\ifanon\anonfalse
\ifanon
  \usepackage{neurips_2024}            
\else
  \usepackage[preprint]{neurips_2024}  
\fi

\makeatletter\renewcommand{\@noticestring}{Preprint.}\makeatother

\usepackage[utf8]{inputenc}
\usepackage[T1]{fontenc}
\usepackage[hidelinks]{hyperref}
\usepackage{xurl}
\usepackage{booktabs}
\usepackage{amsfonts}
\usepackage{amsmath}
\usepackage{amssymb}
\usepackage{nicefrac}
\usepackage{microtype}
\usepackage{graphicx}
\usepackage{xcolor}

\graphicspath{{figures/}}

\title{Memory in the Loop: In-Process Retrieval as\\ Extended Working Memory for Language Agents}

\author{%
  Yusuf Khan \\
  \texttt{yusuf@mykhan.me}
  \And
  Carlo Lipizzi \\
  \texttt{clipizzi@stevens.edu}
}

\begin{document}
\maketitle

\begin{abstract}
Language agents run a loop of observing, reasoning, and acting, but the memory they reason
over is treated as something \emph{outside} that loop: a store queried at most once per
turn. We study the regime in which memory moves \emph{inside} the loop, read and written on
every reasoning step. The standing objection is latency: networked vector stores answer in
tens to hundreds of milliseconds. That cost is a property of \emph{where the store lives},
not of the pattern; an in-process store answers in ${\sim}100\,\mu\mathrm{s}$, and at that
speed the per-step tax collapses. By the extended-mind parity criteria, a store fast enough
to be transparent to a reasoning step, in a loop wired to consult it, functions as extended
working memory. The claim is \emph{causal}, not correlational: holding the per-turn memory
budget fixed and varying only the store's answer speed, redundant actions rise
monotonically from $0.0$ of $12$ at in-process speed to $7.2$ of $12$ at a $110\,$ms round
trip (two models, exact permutation $p{=}0.0079$). End to end, a bounded context window
drives recall to $0/5$ across four GPT-5-class models; in-loop memory recovers
$3.6$--$4.8/5$ with its misses all traced to the agent's read policy, never the store, and a
write-side dedup gate, fixed in advance from that diagnosis, recovers $4.8$--$5.0/5$. An
instructed restate-every-reply baseline also solves the task, but pays per-turn rent: by
turn 25 the gated store is cheaper on every model at list prices, at equal judged accuracy.
The remaining bottleneck is network embedding (${\sim}200$--$400\,$ms); a small local
embedder returns the complete operation to a measured ${\sim}40\,\mu\mathrm{s}$.
\end{abstract}

\section{Introduction}
Language agents are defined by a loop: observe, reason, act, repeat. Yet the memory they
reason over is typically treated as something \emph{outside} that loop---a database the agent
queries once per turn and otherwise leaves alone. This paper asks what happens when memory
moves \emph{inside} the loop: when an agent can read and write an associative store on
\textbf{every step} of its reasoning, as cheaply as it accesses its own context window.

We call this regime \textbf{memory in the loop}, echoing---and extending---the familiar
``human in the loop.'' The obstacle has always been latency. A networked vector store answers
in $50$--$200\,$ms (\S\ref{sec:background}); an agent that consults it at every step pays that
cost repeatedly, and recent work shows in-loop retrieval can inflate end-to-end latency by up
to $83\times$~\citep{searchagenteff2025}. The field has answered on two fronts. Systems work
keeps retrieval in the loop and hides its cost at the serving layer:
SearchAgent-X~\citep{searchagenteff2025} schedules requests by priority and makes retrieval
non-stalling. Industry ``memory-first'' guidance instead moves memory out of the loop, into a
layer queried once at the start of a turn and updated at the end~\citep{mem0loop2026}.

Both answers take the store's latency as given. We argue it is an assumption, not a law. The
latency that makes in-loop
retrieval prohibitive is a property of \emph{where the store lives}, not of the in-loop pattern
itself. An \textbf{in-process} store answers in ${\sim}100\,\mu\mathrm{s}$---three orders of magnitude
below the network regime---and at that speed the amplification collapses: the entire
per-turn in-loop tax measures ${\sim}1.7\,$ms (Table~\ref{tab:tax}). More precisely, the per-turn network tax is
$S\times\text{RTT}$, linear in how often the agent touches memory, which is why a networked
store forces agents to \emph{ration} retrieval; an in-process store drives that tax to ${\sim}0$
and removes the rationing. The tradeoff the efficiency literature treats as fundamental
dissolves; this substrate-level answer complements the serving-layer one rather than opposing
it (\S\ref{sec:thesis}). And the premise is causal, not correlational: holding a per-turn
memory budget fixed and varying only store latency, the task outcome itself flips
(\S\ref{sec:results}).

This is more than an optimization. We ground it in the \textbf{extended-mind}
thesis~\citep{clark1998extended}: an external resource becomes \emph{constitutive} of cognition
only when it is constantly available, directly available without difficulty, and
automatically endorsed upon retrieval. A
$100\,$ms network call is a tool the agent \emph{consults}; a $100\,\mu\mathrm{s}$ in-process store is
\emph{always there}---it clears the latency bar the parity principle sets, and in a loop
wired to consult it becomes genuine extended \textbf{working memory} rather than an external
database (Figure~\ref{fig:parity}). Latency decides whether a store is \emph{eligible} to be
part of the agent's mind; the loop's wiring decides whether it actually is
(\S\ref{sec:parity}).

This is the first step in a broader line of work on memory as a cognitive resource for
language agents. It establishes \emph{when} memory can participate in reasoning: on every
step, once retrieval is cheap. What an agent should retain, and how it should organize what
it retains, are the natural next questions (\S\ref{sec:conclusion}).

\paragraph{Contributions.}
(1)~We lift \emph{retrieval frequency} (per-turn vs.\ per-step) from a serving-layer knob in
RAG systems~\citep{fan2024ragsurvey,hu2025memoryage} to an agent-level design axis, and show
that store latency is what gates it (\S\ref{sec:thesis}).
(2)~We reinterpret the parity principle as an \emph{engineering criterion}---a latency
budget---and argue that in-process latency makes external memory \emph{eligible} as
constitutive working memory, which a loop wired to consult it then realizes
(\S\ref{sec:parity}).
(3)~We show store latency is \emph{causal} to task outcome, not merely to wall-clock
(\S\ref{sec:causalmethod}, \S\ref{sec:causalresults}): under a fixed per-turn memory budget, a scripted
loop guard flips from $0$ to $10/10$ redundant actions, and a real-LLM guard traces a
monotone dose-response---$0.0$ redundant at in-process speed, $1.4$--$1.6$ at $+15\,$ms,
$7.2$ of $12$ at $+110\,$ms where no lookup is affordable (two models, five seeded workloads
per rung, exact permutation $p{=}0.0079$, zero guard errors).
(4)~We demonstrate the regime end-to-end, against the baselines a reviewer would demand:
across all four models of a GPT-5 ladder under a bounded window, recall improves from $0/5$
(every baseline and window-aware run) to $3.6$--$4.8/5$ with in-loop memory, and to
$4.8$--$5.0/5$, $18$ of $20$ runs perfect, once a dedup gate, its threshold fixed in advance, stops duplicate
saves from crowding the fixed read window (same read policy); store ops live
at p50 $80$--$165\,\mu\mathrm{s}$ (\S\ref{sec:results}). An instructed restate-every-reply
baseline reaches $5/5$ on this five-fact task, beating the memory tools here; we show why
restatement stops scaling: it pays a per-turn cost
that grows with the working set, exactly what the store avoids; measured at 25 turns, the
gated store finishes cheaper on every model (\S\ref{sec:results}, Figure~\ref{fig:tokencurve}). In the ungated memory
condition, every miss traces to the agent's read policy, never to the store. The remaining
bottleneck, network embedding, closes to a measured ${\sim}40\,\mu\mathrm{s}$ complete
operation with a small local embedder (\S\ref{sec:embedding}).

\section{Background}
\label{sec:background}
\paragraph{Working memory, from cognition to context windows.}
The Baddeley--Hitch model---a central executive coordinating limited-capacity buffers, later
extended with an episodic buffer~\citep{baddeley1974working,baddeley2000episodic}---is the
standard lens. Applied to language agents, the LLM plays the central executive and the context
window plays the buffer; its capacity limit motivates externalization. Growing the window does
not close the question, for three size-independent reasons. \emph{Rent}: every window token is
reprocessed on every step, so context is paid for repeatedly, while an external store holds
facts at no recurring cost and charges microseconds only for what is fetched. \emph{Findability}: models
reliably miss facts buried in the middle of long contexts~\citep{liu2023lost}, whereas
associative retrieval goes to the fact by meaning. \emph{The cliff}: a long-lived agent
eventually exceeds any window, and the failure is abrupt---our window$\times$facts sweep
(\S\ref{sec:limitations}) locates task failure exactly where the fact span $D$ exceeds the window,
across three task families. The window is the working surface; it cannot also be the archive.
The move has a biological precedent: Ericsson and Kintsch's \emph{long-term working
memory}~\citep{ericsson1995ltwm} shows experts extending working capacity through fast,
reliable retrieval structures in long-term memory---and there, as here, the criterion that
makes the extension work is retrieval speed.

\paragraph{The agent loop and CoALA's internal actions.}
CoALA~\citep{sumers2023coala} frames a language agent as modular memory (working memory plus
episodic/semantic/procedural long-term memory) acted on by three internal actions:
\emph{reasoning} (update working memory), \emph{retrieval} (read long-term memory), and
\emph{learning} (write long-term memory). CoALA fixes retrieval as an action but is silent on
\emph{how often} it can fire; we make that the central variable.

\paragraph{Two latency regimes.}
Retrieval latency clusters into two regimes that differ by orders of magnitude: networked/disk
(cloud vector DBs ${\sim}110\,$ms---a representative cross-region round trip, which we reproduce
against a live Qdrant in \S\ref{sec:results}; RAG $50$--$200\,$ms) and
in-process/in-memory (embedded stores report ${\sim}0.25\,$ms local scans~\citep{lodedb}; our
live store ops measure p50 $80$--$165\,\mu\mathrm{s}$, \S\ref{sec:results}). Around them sit the network
embedding API (${\sim}200$--$400\,$ms, measured in \S\ref{sec:results}) and the LLM reasoning
step itself (${\sim}1\,$s). The gap between these regimes is the substance of the argument:

\begin{center}
\small
\begin{tabular}{lll}
\toprule
Access path & Representative latency & Source \\
\midrule
in-process store op (vxdb) & $80$--$165\,\mu\mathrm{s}$ (p50) & measured live, \S\ref{sec:results} \\
embedded store, local scan & ${\sim}0.25\,$ms & \citep{lodedb} \\
same-host client-server (Qdrant, loopback) & ${\sim}0.35\,$ms p50 & measured, \S\ref{sec:results} \\
cloud vector DB round trip & ${\sim}110\,$ms & modeled + measured, \S\ref{sec:results} \\
network embedding API & ${\sim}200$--$400\,$ms & measured, \S\ref{sec:results}/\S\ref{sec:embedding} \\
one LLM reasoning step & ${\sim}1\,$s & typical, \S\ref{sec:results} wall-clocks \\
\bottomrule
\end{tabular}
\end{center}

\section{The Memory-in-the-Loop Thesis}
\label{sec:thesis}
\begin{figure}[t]
  \centering
  \includegraphics[width=\linewidth]{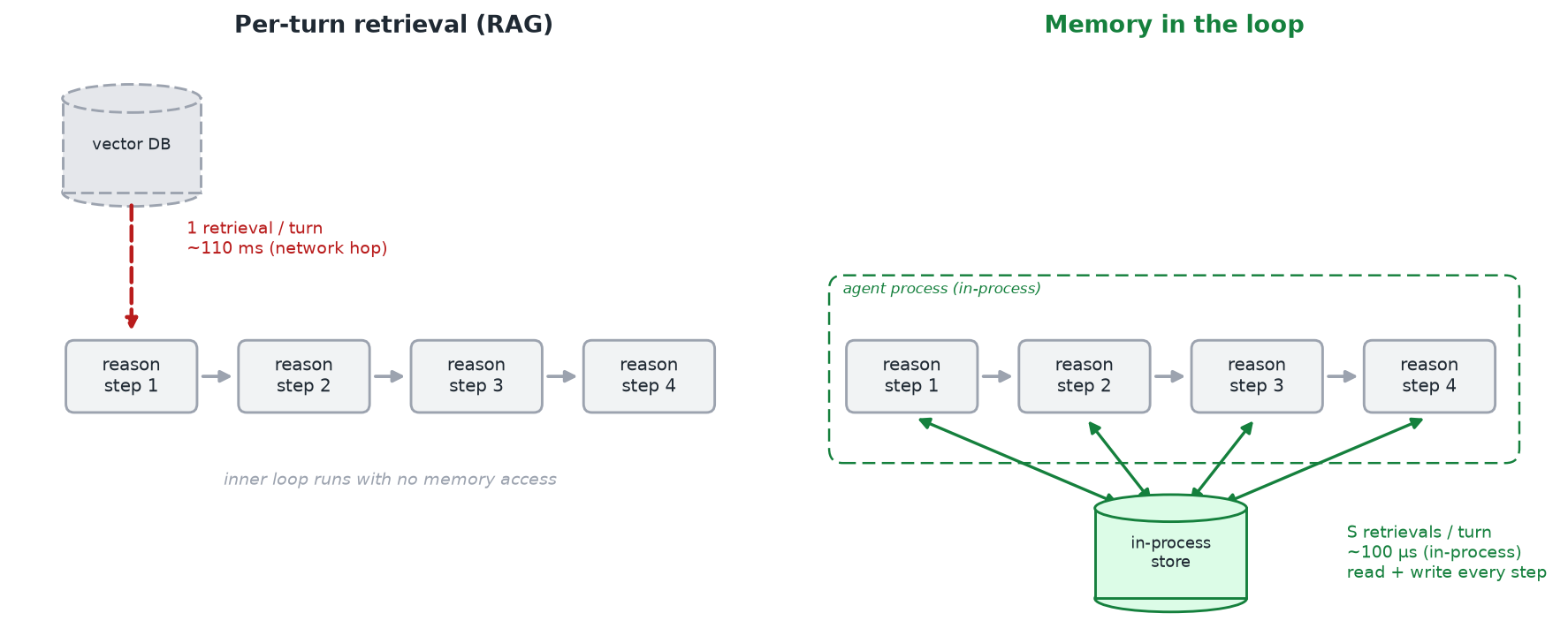}
  \caption{Retrieval frequency. \textbf{Left:} per-turn retrieval (a single network fetch, then
  an inner loop with no memory access). \textbf{Right:} memory in the loop---an in-process store
  read and written on every step.}
  \label{fig:loops}
\end{figure}

\paragraph{Definition (retrieval frequency).}
Let a \emph{turn} be one user/environment interaction and a \emph{step} be one reasoning
iteration within a turn (an LLM call). \emph{Per-turn retrieval} fetches at most once per turn
(classic RAG). \emph{Per-step retrieval} (memory in the loop) may read and write the store at
every step (Figure~\ref{fig:loops}).

\paragraph{The amplification argument and its inversion.}
\citet{searchagenteff2025} show that with in-loop retrieval, end-to-end latency scales with
per-step retrieval latency and is amplified (up to $83\times$) relative to per-turn RAG: each step blocks on
retrieval, and the stalls cascade through the serving layer (evicted KV-cache, missed
scheduling deadlines); \S\ref{sec:results} separates the direct dead time from that
cascade. Their system, SearchAgent-X, keeps retrieval interleaved and
removes the blocking at the serving layer: priority-aware scheduling and non-stall retrieval
over high-recall approximate search. We attack the same term one level down. Their measurements
are taken where $t_\text{store}$ is large ($0.6$--$4.4\,$s of local ANN search), so the
amplification term $S\cdot t_\text{store}$ is large. Reduce $t_\text{store}$ from ${\sim}10^2$--$10^3\,$ms
to ${\sim}10^{-1}\,$ms (in-process) and the term becomes negligible with no scheduling machinery
at all. Both fixes keep memory inside the loop; they compose. The prescription we reject is the
industry ``memory-first'' pattern that moves memory out of the loop into a service queried once
per turn~\citep{mem0loop2026}.

\paragraph{Cost model.}
End-to-end latency is approximately
\begin{equation}\label{eq:cost}
\text{E2E} \;\approx\; \sum_{\text{steps}} \bigl(t_\text{reason} + f\cdot(t_\text{embed} + t_\text{store})\bigr),
\end{equation}
where $f$ is the per-step retrieval frequency. In-loop memory is viable when
$f\cdot(t_\text{embed} + t_\text{store}) \ll t_\text{reason}$. \S\ref{sec:results} measures
$t_\text{store}\approx 100\,\mu\mathrm{s}$ (in-process) vs.\ $t_\text{store}+\text{RTT}$ (network), and
confirms $t_\text{embed}$ (${\sim}200$--$400\,$ms over the network) is the dominant residual
term.

\begin{figure}[t]
  \centering
  \includegraphics[width=0.9\linewidth]{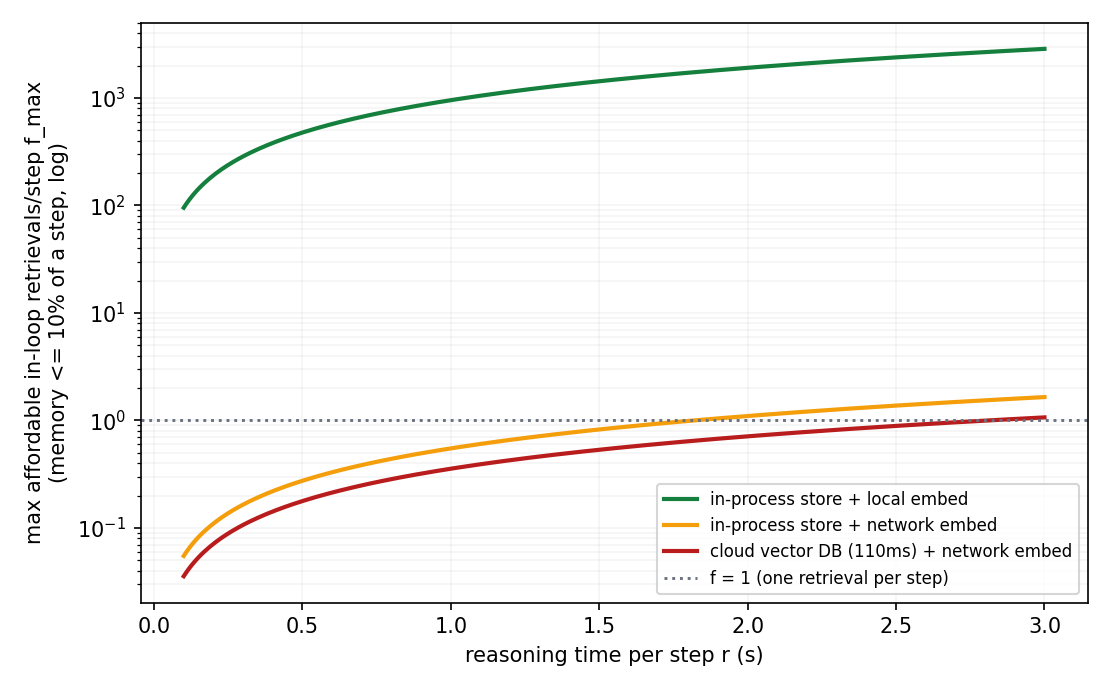}
  \caption{The feasibility frontier: affordable in-loop retrievals per step ($f_{\max}$,
  log scale) against per-step reasoning time, for three memory stacks under a $10\%$ latency
  budget. The dotted $f{=}1$ line is the cliff below which an agent cannot afford a single
  per-step lookup; the in-process$+$local-embedder stack sits three orders of magnitude
  above it, while both network stacks sit below it at ${\sim}1\,$s reasoning steps (the
  network-embedder stack crosses $f{=}1$ only near $r\approx1.8\,$s).}
  \label{fig:frontier}
\end{figure}

\paragraph{Feasibility frontier.}
Fix a budget share $\beta$: memory may add at most a fraction $\beta$ of end-to-end time. With
per-retrieval cost $c = t_\text{embed} + t_\text{store}$ and per-step reasoning time $r$
(Equation~\ref{eq:cost}'s $t_\text{reason}$), the affordable per-step frequency is
\begin{equation}\label{eq:frontier}
f_{\max} \;=\; \frac{\beta}{1-\beta}\cdot\frac{r}{c}.
\end{equation}
Measured, at $\beta{=}0.1$ and $r{=}1\,$s: an in-process store with a local embedder
($c\approx116.6\,\mu\mathrm{s}$) affords $f_{\max}\approx\mathbf{953}$ retrievals per step; the same
store behind a network embedder ($c\approx202\,$ms) affords $\mathbf{0.55}$; a cloud vector store plus
network embedder ($c\approx312\,$ms) affords $\mathbf{0.36}$. Below $1$, an agent cannot
afford even one lookup per step---rationing is forced arithmetic, not a design taste. Near
$10^3$, per-step access is effectively free (Figure~\ref{fig:frontier}).

\paragraph{What per-step memory unlocks.}
Once a read/write is effectively free relative to an LLM step, capabilities become practical
that per-turn retrieval cannot afford: (i)~\emph{working memory}---offload findings and recall
the relevant ones at each step, decoupling effective context from window size;
(ii)~\emph{deduplication/novelty}---check every observation against what has been seen;
(iii)~\emph{loop/action guards}---check before each action whether an equivalent one was already
taken; (iv)~\emph{per-step grounding}---verify claims against retrieved support as produced.

\section{Memory as Constitutive Cognition: the Parity Principle as an Engineering Criterion}
\label{sec:parity}
The engineering contribution here does not depend on any philosophical reading. The latency
measurements, the feasibility frontier, and the causal results stand on their own. We invoke
the extended-mind thesis not as a metaphor but as motivation: if cognition runs on the
interaction between reasoning and continuously available memory, the long separation between
language models and external memory looks like a technological limit rather than a cognitive
necessity. Set the philosophy aside and the engineering result is unchanged.

The extended-mind thesis~\citep{clark1998extended} holds that an external resource can be a
\emph{constitutive part} of a cognitive process, not merely an input to it. Its criteria are
the ones that make Otto's notebook part of Otto's memory rather than a reference he consults:
the notebook is ``a constant in Otto's life,'' its information is ``directly available
without difficulty,'' and upon retrieving information ``he automatically endorses
it''~\citep{clark1998extended} (a fourth criterion, past conscious endorsement, they float
only tentatively).

\begin{figure}[t]
  \centering
  \includegraphics[width=0.92\linewidth]{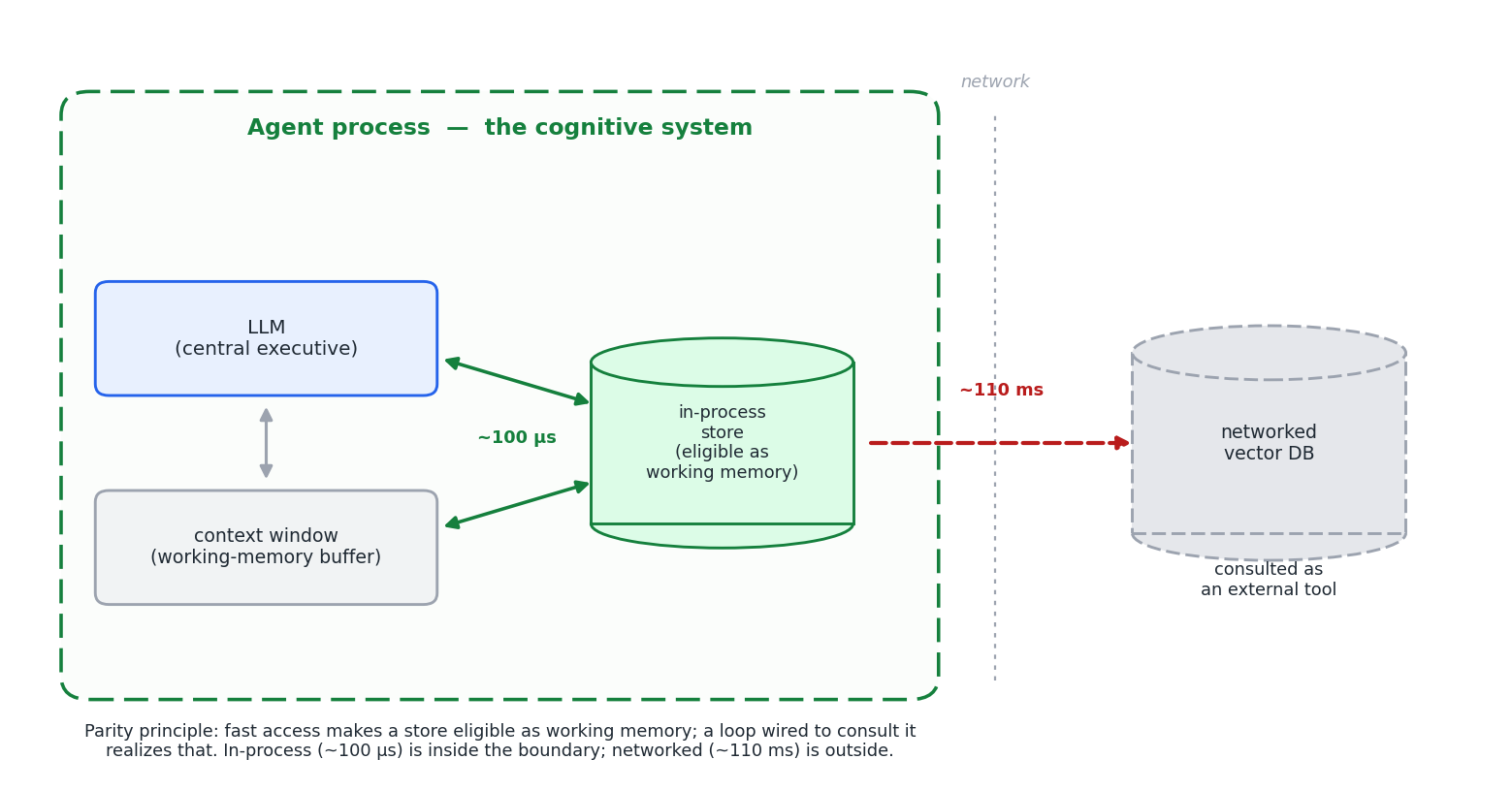}
  \caption{Latency as the cognitive boundary. An in-process store (${\sim}100\,\mu\mathrm{s}$) sits
  inside the agent's process, where its speed makes it \emph{eligible} as extended working
  memory that a loop wired to consult it realizes; a networked store (${\sim}110\,$ms) is
  consulted across the boundary as an external tool.}
  \label{fig:parity}
\end{figure}

We read the first two criteria---constancy and direct availability---as a \textbf{latency
budget}: they are satisfied to the degree that access is fast enough to be transparent to the
ongoing process. For an agent, an in-process store at ${\sim}100\,\mu\mathrm{s}$ is below the threshold
of a reasoning step and is effectively transparent; a networked store at ${\sim}100\,$ms
interrupts the step and is consulted as an external service (Figure~\ref{fig:parity}). The
third criterion, automatic endorsement, is a property of the \emph{loop} rather than the
store, and we return to it below. This
yields the paper's central claim, stated as an engineering criterion rather than a metaphor:

\begin{quote}
\emph{An external memory is eligible as constitutive working memory for an agent to the
extent that a read/write is cheap relative to a reasoning step, and becomes so when the loop
is wired to consult it. Latency is what moves a store across the parity threshold---from a
database the agent connects to, into memory the agent has.}
\end{quote}

\paragraph{Endorsement is the loop's job, not the store's.} Cheap access is necessary, not
sufficient. Our own transcripts supply the counterexample: the agents of \S\ref{sec:results}
write to their $100\,\mu\mathrm{s}$ stores throughout the task (up to $17$ items) yet read
them exactly once per run, through a fixed $k{=}8$ window---full availability, thin
coupling---and every ungated miss in Table~\ref{tab:matrix} is a fact the store held that this
single bounded read never surfaced. Clark and Chalmers' third criterion fails there, and it
fails in the loop's wiring, not in the store's latency. The loop guard of
\S\ref{sec:causalmethod} satisfies endorsement \emph{by construction}: scaffold code consults
the store before every action, so retrieval happens and its answer is acted on, on every
step. The division of the criteria is precise: constancy and direct availability
(criteria~1--2) are latency properties a store either has or lacks, but Clark and Chalmers'
criterion~1 also carries a usage clause---Otto ``rarely take[s] action without consulting''
the notebook---which, like automatic endorsement (criterion~3), is a wiring property the
loop either implements or leaves to the model's tool-calling disposition. Latency buys
availability; the loop's wiring buys consultation and endorsement.

The gated arm of \S\ref{sec:method} adds a second instance of wiring-by-construction, on
the write side: \texttt{remember} consults the store before every save (one $k{=}1$ read,
scaffold code, invisible to the model), the same pattern the guard applies to actions. Its
near-ceiling scores (Table~\ref{tab:matrix}) are therefore not a counterexample to the
endorsement criterion but a measurement of half of it: consultation is wired into the
write path and delivers, while read-side endorsement still rests on the model's
disposition (one ask-triggered recall per run). The trip store, gated or not, is
accordingly not claimed as extended working memory; the loop guard remains the arm where
the loop itself meets the criteria by construction.

\paragraph{The threshold is indexed to the step, and that is the claim.} ``Cheap relative to
a reasoning step'' makes the boundary of working memory relative to the cognizer's cycle
time, and we take that horn deliberately. Working memory has always named a role relative to
a processing cycle, not an absolute latency band. Run a human through
Equation~\ref{eq:frontier}: with deliberate acts near $r{\approx}1\,$s and access costs of $50$--$200\,$ms,
$f_{\max}$ at $\beta{=}0.1$ lands at ${\approx}0.6$--$2$---right at the $f{=}1$ cliff, which
matches the psychological facts: human working-memory access is rationed, capacity-limited,
and attention-gated. The in-process store sits three orders of magnitude above the same
cliff for an LLM agent. One criterion classifies both cognizers correctly, and it predicts
that a $110\,$ms store \emph{is} working memory for a slow enough cognizer---a prediction we
accept, since Figure~\ref{fig:frontier} plots exactly this dependence on $r$. For a
cognizer at $r \approx 1$\,s the verdict is graded, not binary: a $110$\,ms store lands
cliff-marginal ($f_{\max} \approx 1$), in the same band the equation assigns to human
working-memory access, and is excluded decisively only where the per-turn budget sits
below a single round trip (\S\ref{sec:causalmethod}). Two checks
guard against over-reading the $\beta{=}0.1$ point. The ordering is $\beta$-invariant:
$f_{\max}$ carries the common factor $\beta/(1-\beta)$ for every stack, so the
human-at-the-cliff, in-process-far-above classification holds at any budget share, not only
a tenth. And the same criterion places Otto's own notebook correctly---paper access on the
order of seconds sits below the $f{=}1$ cliff at any $\beta\le 1/2$ (there
$f_{\max}\le r/c<1$, since consulting a page runs slower than a reasoning step; a larger
$\beta$ would mean spending most of each step on memory access, which no working-memory
reading permits), so our latency reading calls
the notebook extended \emph{long-term} memory, not working memory, matching Clark and
Chalmers' treatment of it as a standing, non-occurrent belief.

\paragraph{Relation to the Library Theorem.} The closest prior treatment of extended
cognition for tool-using agents, Mainen's Library Theorem~\citep{mainen2026library},
endorses the parity principle for external stores and, in the same section, rejects the
working-memory analogy for context windows: ``the mechanisms differ too deeply for that
comparison to carry weight.'' We concede the mechanism disanalogy in full---no attentional
gating, no rehearsal, no phonological loop is claimed here. ``Working memory'' in this paper
names the functional slot in the agent architecture (the sense the field already uses, as in
CoALA~\citep{sumers2023coala}), not the human mechanism. The two accounts then compose
rather than compete: Mainen prices access in \emph{page reads} and makes organization the
gate; we price it in \emph{wall-clock} and make latency the gate; and his discipline
condition---the store must be authoritative, actually consulted rather than bypassed by a
model that ``already knows''---is the endorsement requirement above.

\paragraph{The coupling-constitution objection.} Adams and Aizawa's classic
objection~\citep{adams2001bounds} is that no measurement of coupling---speed, reliability,
tightness---can show that a resource \emph{constitutes} part of a cognitive process rather
than merely feeding it. We do not claim otherwise. The claim here is functional role parity
at the step level: at the grain of description where the field already counts a context
window as the agent's working memory, a store read and written within the same step, at a
cost invisible at that grain, occupies the same functional slot. The claim is conditional on
that functionalist reading, which the agent-memory literature presupposes whenever it uses
the term. A reader who rejects the extended-mind gloss wholesale loses nothing that is
measured: every latency number, the feasibility frontier, and the causal results of
\S\ref{sec:causalresults} stand on their own as engineering.

Prior work invokes extended cognition \emph{descriptively}; we use it \emph{prescriptively}:
parity implies a latency target, and, on the functionalist reading the field already uses, meeting it changes the agent's cognitive boundary, not
merely its benchmark scores. The claim is functional, not phenomenal, throughout.

\section{Method}
\label{sec:method}
We instantiate in-loop memory as an ephemeral, in-process semantic store and measure its effect
on a multi-turn agent under bounded context. The reproduction artifact---Docker build,
experiment scripts, and per-run JSON records behind every number in this paper---is available
from the author and accompanies the public release.

\paragraph{Primitive.} \texttt{WorkingMemory} wraps an in-memory vector store:
\texttt{add}, \texttt{recall(query,k)}, \texttt{match(text,threshold)}, and \texttt{close}.
It is \emph{allocated}, not connected to. The store op is instrumented to record per-call
latency, separated from embedding latency.

\paragraph{Task and context pressure.} A six-turn trip-planning conversation in which the user
states five durable constraints early, then asks the agent to list them all. A \emph{windowed
session} caps visible history to the last $w$ messages ($w{=}4$), so early constraints scroll
out---a controlled simulation of finite context. The choice of $w$ is inside the failure
region by design: the window sweep of \S\ref{sec:limitations} scans
$W\in\{1,2,3,4,6\}$ against the fact span $D$ and locates the boundary at $D>W$
(Figure~\ref{fig:sweep}). (The sweep plants one fact per turn, so $D$ counts facts and turns alike;
$w{=}4$ messages is two user--assistant exchanges, and the trip task plants its five
constraints across its first three user turns, a three-turn span against an effective
two-turn window, inside the failure region.) Note the baseline scoring $0/5$ under this design measures the
designed floor---the constraints leave the visible window before the recall
question---rather than model capability; the informed baselines below exist to measure what
prompting alone recovers.

\paragraph{Conditions.} Five conditions, same window, same conversation. \emph{Baseline}: no
memory tools, told nothing about the window. \emph{Aware}: no tools, told plainly that its
visible history is short and earlier messages disappear. \emph{Notes}: no tools, additionally
instructed to end every reply with a \textsc{notes} line restating every constraint so
far---the strongest no-tools strategy, restatement made explicit. \emph{Memory}: the same
window plus \texttt{remember}/\texttt{recall} tools backed by an ephemeral
\texttt{WorkingMemory}; the agent decides when to call them (real tool-calling, OpenAI Agents
SDK). The memory condition's instructions include recall coaching (store every durable fact,
reproduce every recalled item). Its read trigger is ask-only: recall fires when the user
requests earlier information (``When the user asks you to recall or list earlier
information, call \texttt{recall} first'') and never as a self-initiated check before
answering, so the memory scores of \S\ref{sec:results} are a lower bound tied to that
policy, not the mechanism's ceiling. \emph{Memory+gate}: identical to \emph{memory} except
that \texttt{remember} consults the store before writing (one recall, $k{=}1$) and silently
skips the save when similarity $\geq 0.70$, a check equivalent to the store's own dedup
primitives (\texttt{match}/\texttt{seen}) wired into the write path. The tool answers ``saved'' either
way, so the conversation the model sees is unchanged; the store's contents are the only
intervention. The threshold was fixed before the run by an offline replay of the twenty
logged memory runs (\S\ref{sec:results}), and this arm ran after the other cells, as a
direct test of the miss diagnosis (replication note in \S\ref{sec:limitations}).
The informed baselines exist precisely so the tools are
compared against informed prompting rather than against ignorance. Full prompts ship in the
artifact.

\paragraph{Models.} A GPT-5 ladder, small $\rightarrow$ latest: \texttt{gpt-5-nano},
\texttt{gpt-5-mini}, \texttt{gpt-5}, \texttt{gpt-5.5}.

\paragraph{Metrics.} Task success (constraints recalled / 5), reported as mean with min--max
range over five \emph{repeats} per model per condition---independent re-samples of the fixed
conversation at the provider's default sampling; the harness exposes no seed parameter. A
second, rubric-based LLM grade (\texttt{gpt-5-mini}) runs alongside the keyword grader;
per-operation store latency (p50), pooled across repeats, treating operations as
exchangeable; percentiles use the harness's nearest-rank convention (the sorted value at
zero-based index $\lfloor n/2\rfloor$ for the median), so reported medians are the upper of the two
central values at even $n$; embedding latency; wall-clock; token counts. For the causal
experiments we report exact permutation $p$-values for the pre-designated primary contrasts
(two-sided, statistic the difference of arm means, all $\binom{10}{5}{=}252$ partitions
enumerated; \S\ref{sec:causalresults}).

\paragraph{A 25-turn extension: the rent measured on the turns axis.} The same task
stretched to 25 turns: the same three constraint-planting turns, the same final ask, and
21 filler turns in between (trip chatter and distractors that state no new durable user
facts and avoid every constraint keyword, so grading stays uncontaminated). Conditions
baseline, notes, and memory+gate, five repeats per model, token usage logged per turn.
All three conditions ran in one sitting; results in \S\ref{sec:results}
(Figure~\ref{fig:tokencurve}).

\subsection{Causal test: the loop-guard task}
\label{sec:causalmethod}
The trip task retrieves once per answer, so its outcome measures the value of \emph{having}
memory, not the speed of \emph{reaching} it: it would score identically against a $100\,$ms
store. To test the thesis itself---that store latency, everything else held fixed, changes
task outcomes---we use a task that wants per-step memory: a \textbf{loop guard}. An agent works
through a stream of candidate actions of which a known fraction are semantic duplicates of
earlier ones; before executing each action it may ask the store ``have I already done this?''
(one real \texttt{match} per step). Memory operates under a fixed \textbf{per-turn latency
budget} $B$; each lookup spends its true cost against the budget (lookups only---the
guard's writes after an executed action are never charged, identically in every arm), and
when the remaining budget cannot cover a conservative estimate of the next lookup's cost
(its injected delay plus a pad of three measured store ops and $1\,$ms), the step runs
unguarded. The charged costs are measured wall times, and a nominal $30\,$ms delay lands
at $35$--$41\,$ms inside the container; that overshoot, not the pad, is why the $+30\,$ms
rung of Table~\ref{tab:llmguard} affords $8$--$10$ lookups per run across seeds (per-model
means $8.4$ and $9.2$, the table's $8$--$9$) rather than the naive $12$. $B$ expresses a designer's cap on
memory-added dead time per turn; our operating point $B{=}100\,$ms is $10\%$ of a one-second
interactive step, just under the $\beta{=}0.1$ budget of \S\ref{sec:thesis} (which
allows $\beta/(1{-}\beta)\cdot r = 111\,$ms), and coincides with the
classic ${\sim}0.1\,$s threshold below which interface feedback reads as instantaneous to a
human~\citep{miller1968response,nielsen1993usability}. Because any such choice
is contestable we \emph{sweep} $B$ rather than assert it. The design varies the store's
answer speed and reads off one outcome, redundant actions executed---so the claim is
conditional and explicit: \emph{under a memory budget below one store round trip, latency
alone determines whether the guard exists at all}. One distinction deserves emphasis:
latency in this design never alters what a lookup \emph{returns}---the store's answers are
identical, and correct, in every arm. It alters only whether a lookup \emph{fits the
budget}. The causal path from latency to task damage therefore runs entirely through checks
that never happen, never through answers that go wrong; the ``mistakes'' counted in
Table~\ref{tab:llmguard} are repeated \emph{actions}, not retrieval errors.

\paragraph{Deterministic version (mechanism demonstration).} A scripted agent, no LLM
anywhere: $S{=}20$ candidate actions per turn, half exact duplicates. Guards afforded per
turn follow from the budget by construction ($\lfloor B/\text{op}\rfloor$, capped at $S$),
executed against a real store so guard correctness is real rather than assumed. We present
this as arithmetic made visible, not as an empirical finding: it shows the mechanism, and the
LLM version carries the empirical weight. Latency rungs: the measured in-process op plus
injected round trips of $1$, $5$, $30$, $110$, and $200\,$ms; every cell is averaged over
$25$ independent workload shuffles, and the full budget grid
$B\in\{25,50,100,200,400\}\,$ms is reported alongside the $B{=}100$ operating point.

\paragraph{LLM version.} A real model produces the decisions. The agent triages $24$ security
alerts across $4$ turns ($6$ per turn); half are template-reworded duplicates of earlier
alerts. The model sees only the last $w{=}4$ exchanges, so it genuinely
forgets earlier work; for each alert it answers \textsc{investigate} or
\textsc{skip}-as-duplicate, and a code guard checks the store before any \textsc{investigate}
executes. (Here $w$ counts exchanges---eight messages---where the trip task's $w$ counts
messages; each task's window is stated in its own unit.) The guard sees only what the
system could see in production---never the workload
generator's ground-truth fields. It matches the model's \emph{own} extracted action line
(``\textsc{investigate} host=$h$ sig=$s$'') through the paper's local semantic embedder
(\texttt{potion-retrieval-32M}, \S\ref{sec:embedding}), with the match threshold calibrated
offline on twenty held-out workload seeds disjoint from the experiment's ($F_1{=}1.0$ at
threshold $0.96$ on held-out data). The calibration also documents why the guard does not
match raw alert prose: under a static embedder, same-template alerts about \emph{different}
incidents score higher ($0.94$ max) than true rewordings of the \emph{same} incident ($0.58$
median)---template wording dominates the embedding. Matching the model's extracted action line is the design that survives
that measured failure. On these action lines, true duplicates are near-identical strings (calibration
cosine ${\approx}1.0$ against a worst impostor at $0.918$), so the threshold in effect
performs exact matching on the model's extracted output; the embedder's remaining work is
rejecting same-format lines about different incidents. Scoring uses the \emph{executed set}: an execution is redundant only if the
same host+signature pair was already executed, a guard match that blocks genuinely new work
is counted against the guard (\texttt{guard\_blocked\_new}), and guard precision/recall are
reported with per-lookup records in the artifact. The guard embeds each action line with the
in-process local embedder (\texttt{potion-retrieval-32M}, ${\sim}32\,\mu\mathrm{s}$,
\S\ref{sec:embedding}), and the causal result is stated for that stack: a network embedder
(${\sim}200$--$400\,$ms) would exceed the $100\,$ms budget in \emph{every} arm, arm~(a)
included, which is the feasibility frontier's point, not a confound---the embedder is held
identical across arms, so only the injected store latency varies.

\paragraph{Arms.} Same model, same seeded workloads; prompts are identical between arms
(a) and (b), and arm (c) differs only by its pinned memory dump. \textbf{(a) in-process}: the
guard runs at measured in-process speed, affording every check. \textbf{(b) delayed ladder}:
the identical guard pays an injected $+\{5,15,30,50,110\}\,$ms per lookup---each affordable
lookup genuinely sleeps its delay---so affordability degrades continuously and the outcome
traces a \emph{dose-response} curve rather than a single cliff. \textbf{(c) per-turn RAG}:
the industry-default pattern---store contents retrieved once at turn start and pinned in
context, updates at turn end, no in-loop guard. A budget sweep
($B\in\{100,150,250,500\}\,$ms at the $110\,$ms rung; \texttt{gpt-5-nano}) shows what
happens when the budget rises past one round trip. Models: \texttt{gpt-5-nano} and \texttt{gpt-5-mini}; five seeded
workloads per condition. Primary outcome: redundant investigations executed (of $12$
duplicates); pre-designated contrasts are arm (a) vs.\ the $110\,$ms rung and arm (a) vs.\
arm (c), tested by exact permutation.

\paragraph{Measurement hosts.} All live agent numbers (Table~\ref{tab:matrix}), the causal
experiments, and the reproduction artifact come from one host: an Apple M4 ($10$ cores,
$16\,$GB) inside Docker. The original development host produced the same qualitative pattern
throughout; drift between hosts is reported, not smoothed: the Qdrant loopback gap measured
$7.5\times$ (median) on the original host and $4.2\times$ on the M4, same direction; the
deterministic guard experiment's in-process op measured $85\,\mu\mathrm{s}$ in the audit run of
record and $17.5\,\mu\mathrm{s}$ in the current regeneration.

\section{Results}
\label{sec:results}
\begin{figure}[t]
  \centering
  \includegraphics[width=\linewidth]{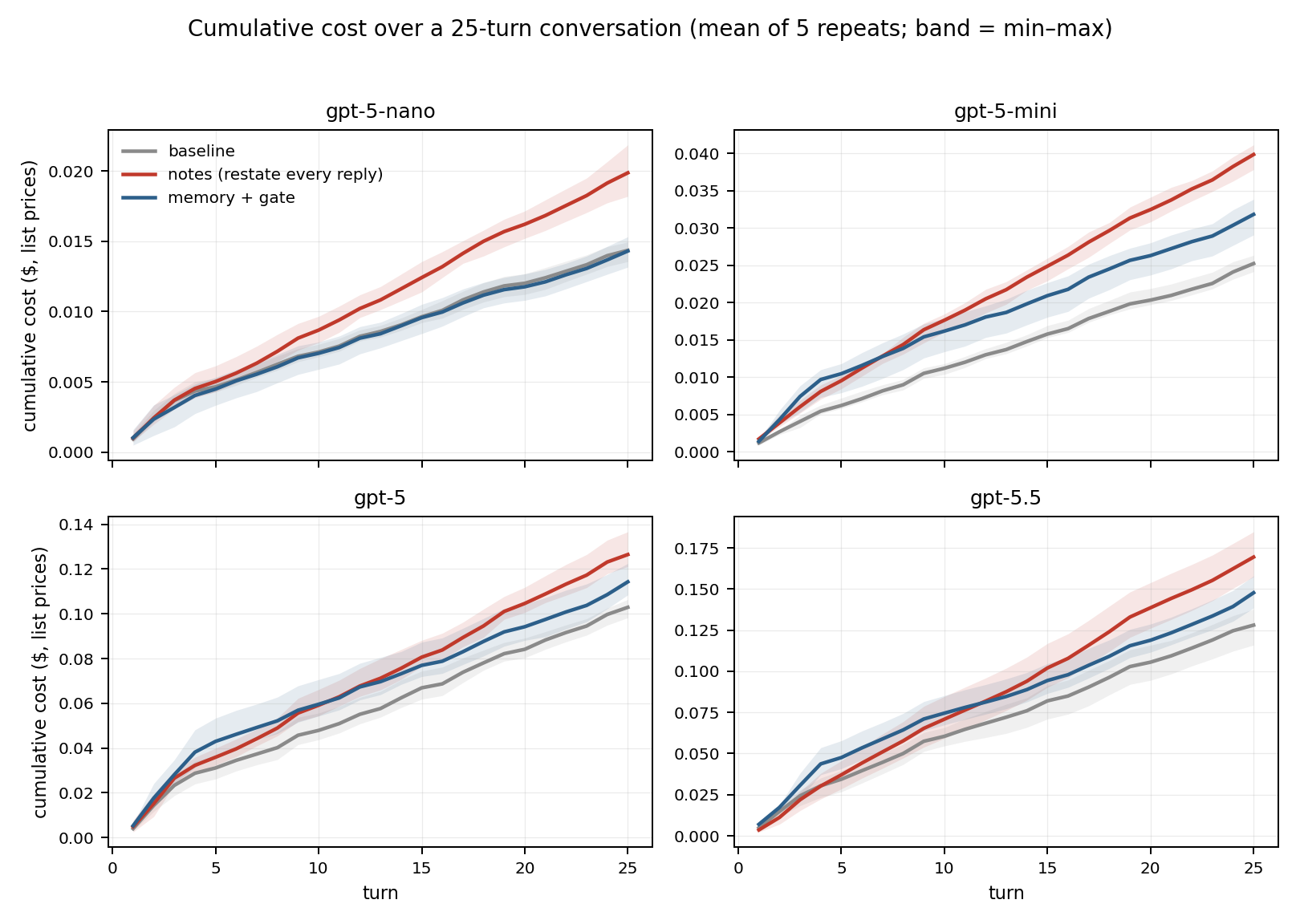}
  \caption{Cumulative dollar cost per turn over the 25-turn conversation (mean of five
  repeats per condition; band = min--max; list prices). Notes climbs every turn;
  memory+gate pays its writes early, then tracks baseline; the curves cross by
  turn 12 on every model.}
  \label{fig:tokencurve}
\end{figure}

\paragraph{The window causes total failure; two mechanisms recover it.}
Over five repeats per model per condition, every baseline run scores $0/5$---and so does
every \emph{aware} run: telling the model its history is windowed changes nothing without a
mechanism to act on it (forty runs, no exceptions). Two mechanisms recover the task
(Table~\ref{tab:matrix}). The \emph{notes} baseline---restate every constraint in every
reply---solves this five-fact task perfectly: $5.0$ in all twenty runs. At five facts and
six turns, disciplined restatement beats memory tools. Its cost is
structural rather than visible here: every live fact is re-emitted in every reply and
re-read on every subsequent step, rent that grows with the working set and is bounded by
what a reply can carry (\S\ref{sec:background}); the token ledger below prices it and the
25-turn extension measures it (by turn 25 the gated store finishes cheaper on every model,
at equal judged accuracy, Figure~\ref{fig:tokencurve}), and the
regime where the store's per-fetch pricing takes over is
exactly the regime this small task never reaches. The \emph{memory} condition reaches
$3.6$--$4.8/5$ with a store that never lost a fact; its misses are read-policy failures,
decomposed below and removed there by a write-side intervention fixed in advance
(\emph{memory+gate}: $4.8$--$5.0/5$). A second, rubric-based LLM grade matches the keyword
grader value-for-value in all one hundred runs. Forgetting is a \emph{context} problem, not a
capability problem: the largest model fails identically to the smallest without a mechanism.

\paragraph{Where the misses live: the read policy, never the store.} The store kept all
$244$ writes across the twenty runs and answered every query it was asked (the pre-fix
runs, in which an id collision under the network embed destroyed $47\%$ of writes, are
archived in the artifact and analyzed in \S\ref{sec:embedding}). The misses have one
shape: \emph{over-writing plus under-reading}. The agents re-save the same constraint in
fresh wordings (up to four copies of a single constraint) and store incidental
non-constraint details besides, growing stores of $7$--$17$ items, then
read exactly once per run through the tool's fixed $k{=}8$ window; near-duplicate copies
of already-covered constraints crowd the eight slots (the median read carries only four
distinct constraints), and a constraint the store demonstrably holds never reaches the reply. The pattern
is one-directional: every run whose store stayed at or under $10$ items scored $5/5$, and
all eighteen misses come from runs whose stores had outgrown the read window ($11$--$17$
items against $k{=}8$), though three larger-store runs still recalled all five: outgrowing
$k$ is necessary for a miss here, not sufficient. The read's contents settle the score
exactly: in all twenty runs, the set of constraints the grader finds in the final reply
equals the set of constraints present in the up-to-eight lines the single recall
returned. Five distinct constraints in the read give $5/5$ (eight runs), four give $4/5$
(six runs), three give $3/5$ (six runs), and duplicate copies of already-covered
constraints occupy two to five of the eight slots in every run. No run drops a constraint
its read surfaced; no run recovers one its read cut off. The over-writing is itself a product of the bounded window: the agent cannot see
its earlier saves, so it saves again. Once a conversation outgrows any context, the
transcript can no longer answer ``did I already store this?''; the store can, and the gate
below operationalizes exactly that check. What stays model-side is variance in what gets
saved at all, which is why we report means with ranges rather than a best run. The store op stays in microseconds live (write p50 $80$--$89\,\mu\mathrm{s}$,
$n{=}44$--$81$ ops per model; the single per-run recall pools to a median of
$165\,\mu\mathrm{s}$ over these $7$--$17$-item stores, $n{=}20$), measured between the
agent's tool calls; the clean-room benchmark is the latency measurement of record
(${\sim}60{,}000$ ephemeral create-fill-destroy lifecycles/sec, recall p50
${\approx}85\,\mu\mathrm{s}$ at $1{,}000$ items, same host).

\begin{table}[t]
  \centering
  \caption{GPT-5 ladder on the windowed recall task: mean constraints recalled (of 5) with
  [min--max] over five repeats per cell. \emph{Baseline} is told nothing about its window;
  \emph{aware} is told the window is short; \emph{notes} is additionally instructed to
  restate all constraints in every reply; \emph{memory} has \texttt{remember}/\texttt{recall}
  tools. \emph{Memory+gate} is the memory condition unchanged except for a silent write-side
  dedup check (\S\ref{sec:method}); it was run after the other cells, as an
  intervention on the miss mechanism diagnosed from them, its threshold fixed in advance
  from their logs. The second grader's mean matches
  every cell shown. Memory means summarize per-repeat
  integers that span the stated ranges (e.g.\ \texttt{gpt-5.5} memory: $5,5,4,3,3$;
  \texttt{gpt-5} memory+gate: $4,5,5,5,5$).}
  \label{tab:matrix}
  \begin{tabular}{lccccc}
    \toprule
    Model & Baseline & Aware & Notes & Memory & Memory+gate \\
    \midrule
    \texttt{gpt-5-nano} & 0.0 [0--0] & 0.0 [0--0] & 5.0 [5--5] & 4.8 [4--5] & 4.8 [4--5] \\
    \texttt{gpt-5-mini} & 0.0 [0--0] & 0.0 [0--0] & 5.0 [5--5] & 3.6 [3--5] & 5.0 [5--5] \\
    \texttt{gpt-5}      & 0.0 [0--0] & 0.0 [0--0] & 5.0 [5--5] & 4.0 [3--5] & 4.8 [4--5] \\
    \texttt{gpt-5.5}    & 0.0 [0--0] & 0.0 [0--0] & 5.0 [5--5] & 4.0 [3--5] & 5.0 [5--5] \\
    \bottomrule
  \end{tabular}
\end{table}

\paragraph{The write-side intervention removes the diagnosed misses.} The diagnosis
above makes a specific prediction: stop the duplicate writes and the fixed read recovers
the task. We tested it with a write-side check equivalent to the dedup primitives the
store already ships (\texttt{match}/\texttt{seen}; \S\ref{sec:method}). First, offline: we replayed the twenty logged save streams
under the runs' own embedder through a simulated gate. The simulator is validated exactly:
with the gate off, its reconstructed top-8 matches the eight actually-returned lines
in all twenty runs. Across all $1{,}384$ constraint-to-constraint comparisons a live gate
would have faced, facts about \emph{different} constraints never exceed $0.496$
similarity, so we fixed the threshold at $0.70$ in advance; the replay predicts $5/5$ in
all twenty runs. Then, live (Table~\ref{tab:matrix}, \emph{memory+gate}): $5/5$ in
eighteen of twenty fresh runs (\texttt{gpt-5-mini} rises from $3.6$ to $5.0$ and
\texttt{gpt-5.5} from $4.0$ to $5.0$ on every repeat); stores shrink from $7$--$17$ items
to $4$--$12$, and the gate blocks $109$ duplicate saves with zero cross-constraint errors:
it never rejected a genuinely new fact. The read-content identity holds in all twenty gated
runs as well. The two residual $4/5$ runs have new, narrower shapes: one run never
attempted to save the heights constraint (a write-side omission no store can repair), and
one saved paraphrase pairs that score below any safe threshold under this embedder
(``Traveler is vegetarian'' vs.\ ``User is vegetarian'') plus incidental facts, leaving a
12-item store whose 8-item read again cut the last constraint. Nothing about reading or
answering changed: the same ask-only policy over an unpolluted store reaches the
restatement baseline's score on two of four models and $4.8$ on the others.

\paragraph{What each mechanism costs: the runs' own token ledger.} Every run logged its
token usage, and the ledger prices \S\ref{sec:background}'s rent. The notes discipline
costs $7$--$40\%$ more total tokens than baseline, and its overhead is dominated by
\emph{output} tokens: text the model writes, priced at $6$--$8\times$ the input rate on
this ladder's published price sheets and never cache-discounted. On the one model that ran
without reasoning tokens the overhead isolates to the restatement text itself:
${\approx}8$ output tokens per fact per reply, a constant that multiplies with the
working set and is paid in every reply whether or not the turn needs the facts. The notes
agents restate every constraint stated so far in every reply, though only the final turn asks,
and by turns 5--6 their replies run $1.3$--$3.9\times$ the length of baseline replies. The
memory condition spends more raw tokens than notes ($1.8$--$2.8\times$ baseline), but on
the other side of the price sheet: each remember or recall is a separate model exchange
that re-reads the visible context (${\approx}340$--$530$ input tokens per call on three of
the four models), so a run with $8$--$18$ tool calls re-buys its own transcript that many
times, almost entirely in \emph{input} tokens, the cheap and cacheable kind:
repeated-prefix text is exactly what provider caches price at a tenth of the input rate, a
discount restatement's output can never take. At list prices the two mechanisms land in
one band at this scale (notes $1.06$--$1.50\times$ baseline in dollars across the ladder;
memory $0.86$--$1.83\times$, the floor being \texttt{gpt-5-nano}, which wrote $4{,}067$
fewer output tokens, mostly reasoning, with tools to offload to), and the band decomposes
asymmetrically: restatement pays
flat-rate rent in expensive tokens on a bill that grows with the fact count; the store
pays per use, in cheap tokens, for a read whose size is fixed at $k$ lines however large
the store grows. Five facts and six turns sit below the crossing. The gate leaves this
ledger essentially unchanged: the duplicate calls still happen and still pay their round
trips; it is the store's contents, and with them the score, that the gate repairs.

\paragraph{The rent compounds with conversation length (Figure~\ref{fig:tokencurve}).}
At 25 turns the window still causes total failure (baseline $0/5$, all twenty runs) and
both mechanisms still recover it: notes at $5/5$ in all twenty, its \textsc{notes} line
surviving 24 consecutive hand-copies without dropping a constraint (restatement does not
break at this scale; it pays), and memory+gate at judge $5.0$ in all twenty, with the
keyword grader at $4/5$ in two \texttt{gpt-5} runs whose replies paraphrase the heights
constraint without the keyword. The ask-only read held at length: exactly one recall per
run in all twenty gated runs, fired at the final ask, over stores of $5$--$12$ items (the
$9$--$12$ ceilings are all \texttt{gpt-5}, whose incidental saves outgrew the $k{=}8$ read
again without costing a point here). The
costs diverge. Notes' cumulative bill climbs every turn, its gap over baseline widening at
every checkpoint on every model, while the gated arm pays its writes early and then tracks
baseline (its replies run shorter than baseline's on all four models). At the same
list prices as above, notes finishes at $1.23$--$1.58\times$ baseline across the ladder
and memory+gate at $1.00$--$1.26\times$, undercutting notes on every model; on the
smallest model the gated arm reaches cost parity with baseline itself. The six-turn tie is
the early segment of two different slopes: the late-conversation per-turn gap is roughly
constant, so the cumulative gap grows linearly with length, and nothing in the mechanism
caps it. The facts axis (more constraints rather than more turns) remains unmeasured
(\S\ref{sec:limitations}).

\begin{figure}[t]
  \centering
  \includegraphics[width=0.82\linewidth]{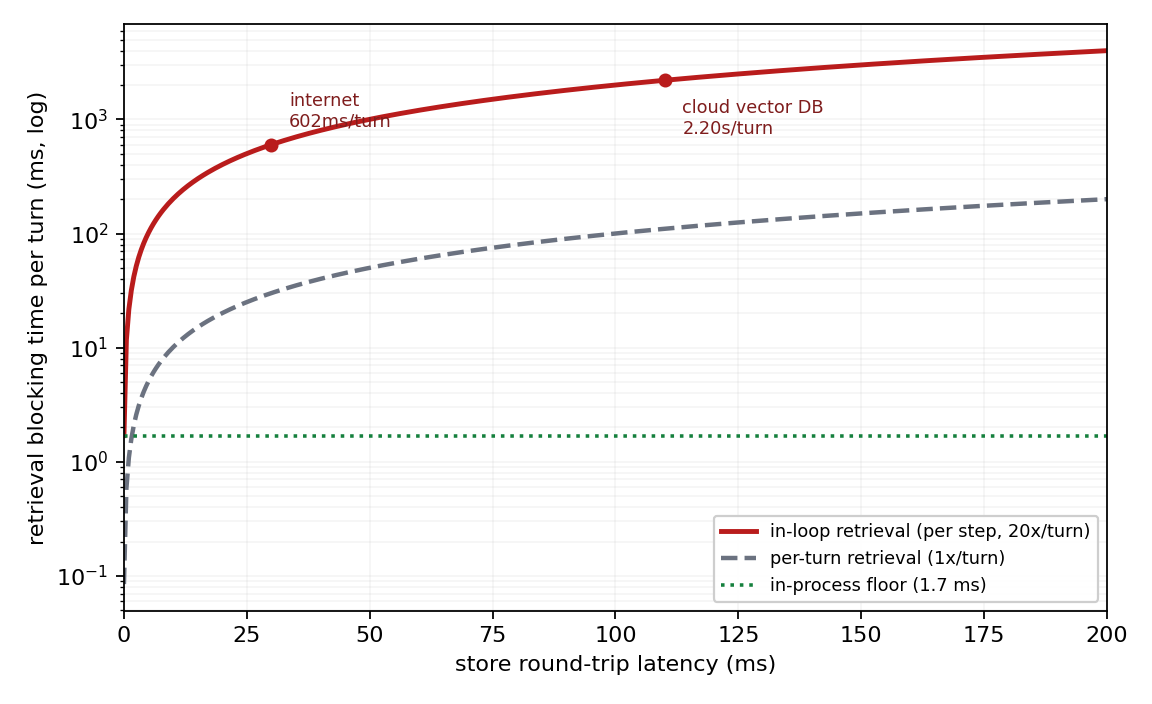}
  \caption{In-loop retrieval tax vs.\ store round-trip latency (log $y$; the measured store op
  of Table~\ref{tab:tax} + modeled network RTT). The per-step curve reaches seconds per turn on
  a network store while the in-process floor stays at ${\sim}1.7\,$ms.}
  \label{fig:tax}
\end{figure}

\paragraph{Reproducing and dissolving the in-loop latency tax.}
We isolate the store op (real vxdb recall, embedding excluded) and model a networked store as
the \emph{same} operation plus a round-trip delay ($\text{network}=\text{store\_work}+\text{RTT}$),
adding only the RTT---charitable to the network store. Let a turn run $S{=}20$ in-loop
retrievals. The retrieval blocking time per turn (Table~\ref{tab:tax}, Figure~\ref{fig:tax}) is
$1.7\,$ms in-process vs.\ $2.20\,$s at a $110\,$ms cloud vector DB---a ${\sim}1{,}300\times$
difference.

\begin{table}[t]
  \centering
  \caption{Retrieval blocking time per turn ($S{=}20$ in-loop retrievals; store op
  ${\approx}85\,\mu\mathrm{s}$ at $1{,}000$ items, measured on the Apple M4 host of
  \S\ref{sec:causalmethod}).}
  \label{tab:tax}
  \begin{tabular}{lcccc}
    \toprule
    Store backend & RTT & per-turn ($1\times$) & per-step ($20\times$) & vs.\ in-process \\
    \midrule
    in-process (vxdb)        & ---     & $0.085\,$ms & $1.7\,$ms  & $1\times$ \\
    same-AZ                  & $1\,$ms & $1.08\,$ms & $21.7\,$ms & $13\times$ \\
    cross-AZ                 & $5\,$ms & $5.08\,$ms & $101.7\,$ms & $60\times$ \\
    internet                 & $30\,$ms & $30.1\,$ms & $602\,$ms  & $354\times$ \\
    cloud vector DB          & $110\,$ms & $110\,$ms & $2.20\,$s  & $1{,}294\times$ \\
    slow                     & $200\,$ms & $200\,$ms & $4.00\,$s  & $2{,}353\times$ \\
    \bottomrule
  \end{tabular}
\end{table}

\paragraph{Live networked baseline (measured).}
To validate the model we push the same $1{,}000$ vectors into vxdb (in-process) and into a real
\textbf{Qdrant} container (gRPC over localhost) and measure per-op recall latency directly
(Table~\ref{tab:live}). Even with \emph{no network}---Qdrant on the same host over loopback---the
client-server boundary costs ${\sim}4.2\times$ at the median ($84.7\rightarrow352\,\mu\mathrm{s}$)
and ${\sim}4.4\times$ at the p99 tail ($94\rightarrow410\,\mu\mathrm{s}$) on this host (the
original development host measured $7.5\times$ and $27\times$; drift reported in
\S\ref{sec:causalmethod}); the tail matters because an in-loop agent pays it
every step. Adding a realistic $110\,$ms WAN to the \emph{measured} Qdrant op reproduces the
modeled $2.20\,$s/turn almost exactly ($2.21\,$s). The loopback number is a hard lower bound on the networked
penalty.

\begin{table}[t]
  \centering
  \caption{Measured per-op recall latency: in-process vxdb vs.\ a real Qdrant container
  (Apple M4 host, $1{,}000$ vectors, $2{,}000$ trials).}
  \label{tab:live}
  \begin{tabular}{lccc}
    \toprule
    Backend & recall p50 & recall p99 & per-step tax ($20\times$/turn) \\
    \midrule
    vxdb in-process (measured)               & $84.7\,\mu\mathrm{s}$  & $94\,\mu\mathrm{s}$ & $1.7\,$ms \\
    Qdrant, loopback (measured)              & $352\,\mu\mathrm{s}$ & $410\,\mu\mathrm{s}$ & $7.0\,$ms \\
    Qdrant $+\,110\,$ms WAN (meas.+proj.)    & $110.4\,$ms  & ---          & $2.21\,$s \\
    \bottomrule
  \end{tabular}
\end{table}

\paragraph{The nuance on ``$83\times$.''}
Whether this dead time dominates \emph{end-to-end} latency depends on how it compares to
reasoning. With per-step reasoning time $r$, the in-loop vs.\ per-turn amplification is
$\frac{S r + S L}{S r + L}$ for store latency $L$ (the full per-retrieval store cost: Equation~\ref{eq:cost}'s
$t_\text{store}$ plus any network RTT). When reasoning dominates ($r\approx 1\,$s),
even a $110\,$ms store yields only ${\sim}1.10\times$ amplification. The formula is
doubly capped, at $S$ and at $1+L/r$, so with $r\approx1\,$s even the expensive
retrieval of~\citet{searchagenteff2025} ($0.6$--$4.4\,$s) could amplify dead time by at
most ${\sim}5.4\times$; their measured $83\times$ is, by their own analysis, a
serving-layer cascade on top of the blocking (retrieval stalls evict KV-cache and miss
scheduling deadlines, forcing recomputation), exactly the machinery their scheduler then
removes. The correct reading is therefore
not ``memory in the loop gives an $83\times$ speedup,'' but: the network tax is $S\times\text{RTT}$,
so a networked store forces \emph{rationing} of retrieval, while an in-process store makes
per-step, per-observation, per-action memory access \emph{feasible at all}---a regime off the
table at network latency.

\subsection{Store latency alone changes the outcome}
\label{sec:causalresults}
Everything above measures time. This section measures \emph{task quality}, with latency as
the only manipulated variable (\S\ref{sec:causalmethod}).

\begin{figure}[t]
  \centering
  \includegraphics[width=\linewidth]{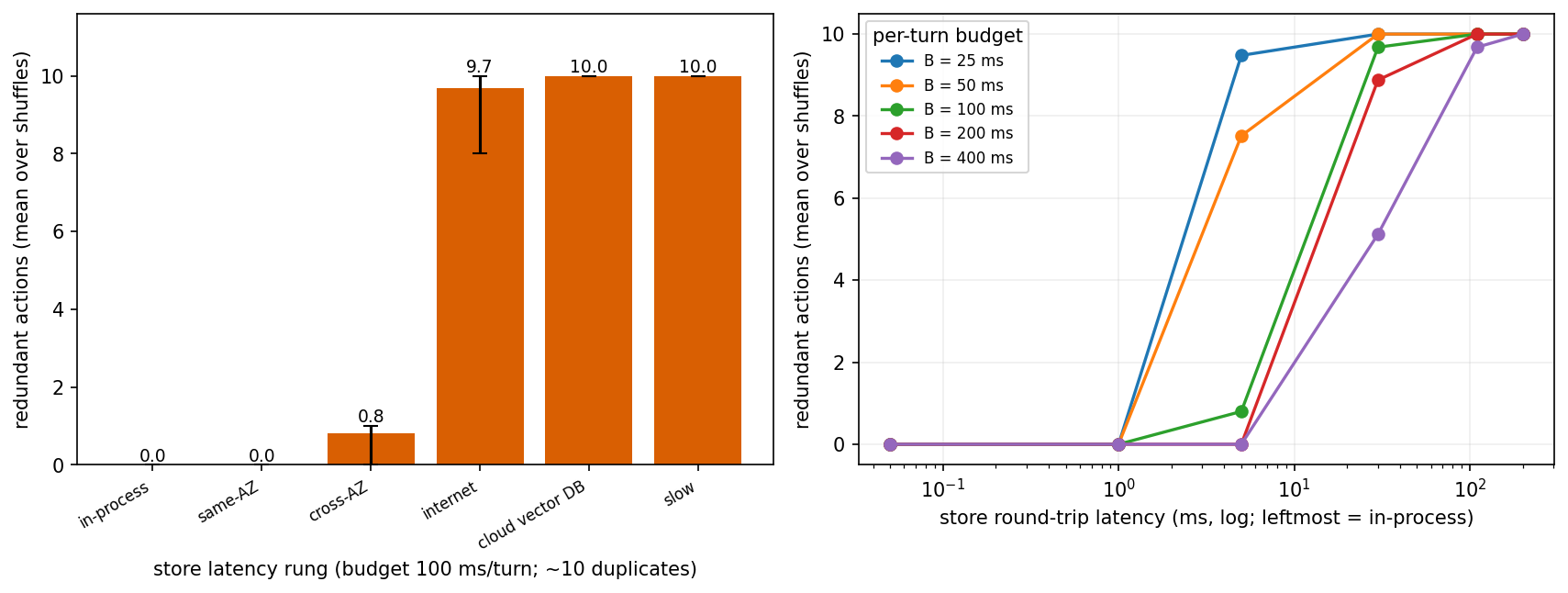}
  \caption{Mechanism demonstration under a fixed memory budget. \textbf{Left:} duplicate
  actions executed per latency rung ($B{=}100\,$ms; mean with min--max over $25$ workload
  shuffles). \textbf{Right:} mean duplicates across the budget $\times$ latency grid; every
  budget has its cliff.}
  \label{fig:causal}
\end{figure}

\paragraph{Deterministic loop guard (mechanism).} With the in-process store (op
${\approx}17.5\,\mu\mathrm{s}$ this run), all $20$ guard checks fit the $100\,$ms budget and
$\mathbf{0}$ duplicates execute, averaged over $25$ workload shuffles. The rungs behave like
a dial: $1\,$ms still affords all $20$ guards; $5\,$ms affords $19$ and leaks $0.8$
[$0$--$1$]; $30\,$ms affords $3$ and leaks $9.7$ [$8$--$10$]; at $110\,$ms not one check
fits and all $\mathbf{10}$ execute. The budget grid (Figure~\ref{fig:causal}, right) runs
the same arithmetic at every $B$ from $25$ to $400\,$ms: the cliff moves, the mechanism does
not. The other horn of the dilemma is paying instead of rationing: guarding every step costs
$0.4\,$ms per turn in-process versus $2.2\,$s at the cloud store. This panel is floor
arithmetic executed against a real store; the empirical weight sits in the LLM runs below.

\begin{figure}[t]
  \centering
  \includegraphics[width=\linewidth]{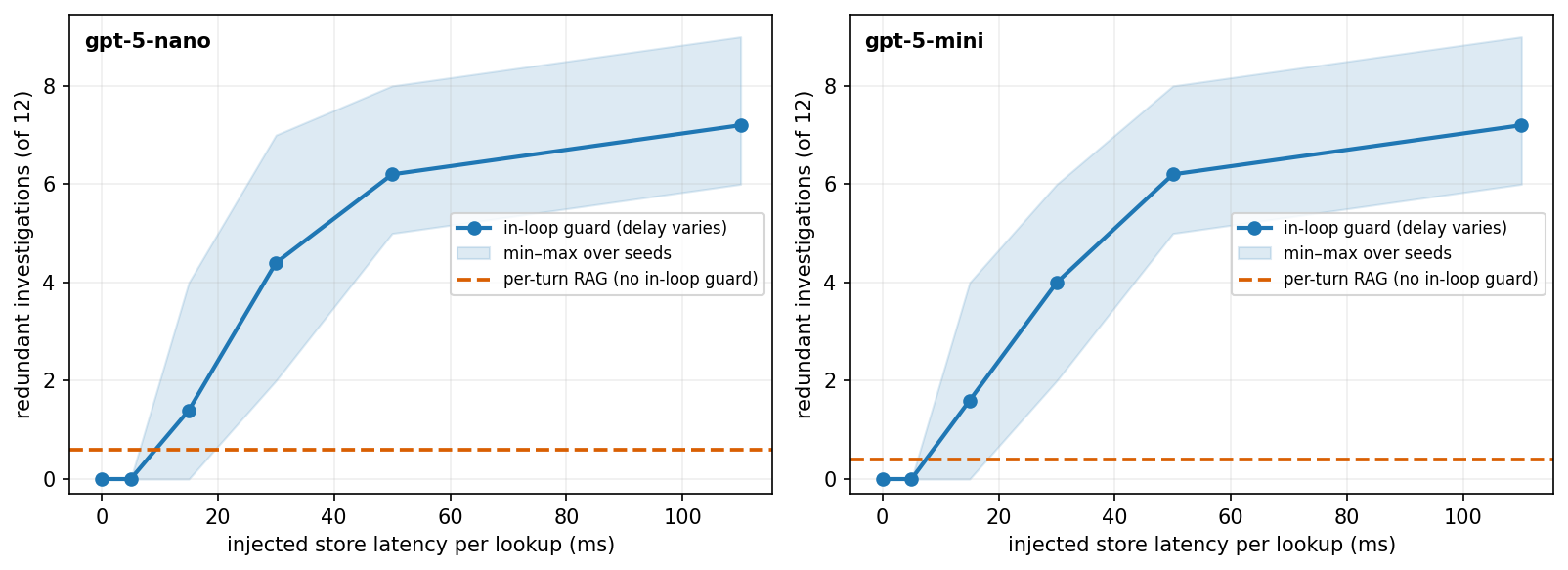}
  \caption{Dose-response: redundant investigations vs.\ injected per-lookup store latency
  under a fixed $100\,$ms/turn budget (mean with min--max band, five seeded workloads;
  dashed line: per-turn RAG reference).}
  \label{fig:llmcausal}
\end{figure}

\paragraph{Real-LLM loop guard: a dose-response.} The same mechanism holds with a real
model producing the actions and a real semantic guard
(Table~\ref{tab:llmguard}, Figure~\ref{fig:llmcausal}). Redundant executions rise
monotonically with injected store latency: $\mathbf{0.0}$ at in-process speed and at
$+5\,$ms (every workload of both models exactly zero), $1.4$--$1.6$ at $+15\,$ms,
$4.0$--$4.4$ at $+30\,$ms, $6.2$ at $+50\,$ms, and $\mathbf{7.2}$ of $12$ at the $110\,$ms
cloud round trip---where not a single lookup fits the $100\,$ms budget, so the guard is
priced out entirely and every duplicate that survived the window executes. Exact permutation
tests on the pre-designated contrast (in-process vs.\ $110\,$ms): $p{=}0.0079$ for both
models, the minimum attainable at five workloads per arm; the $30$ and $50\,$ms rungs also
reach $0.0079$, the $15\,$ms rung $0.048$. Only the two pre-designated contrasts are
confirmatory; the per-rung values are descriptive, and the $15\,$ms rung's $0.048$ would
not survive correction across the twelve tests reported. The guard itself never erred where it ran:
precision and recall $1.0$ wherever defined (a few workloads afford lookups but meet no
true duplicate), and zero new investigations falsely blocked anywhere---matching on the
model's own extracted action lines through the real
embedder, never on oracle fields. The guard does cognitive work the models cannot: they
recognized only ${\sim}5$ of $12$ disguised duplicates from their windowed context; the
store caught the remaining ${\sim}7$ whenever a lookup was affordable.

\begin{table}[t]
  \centering
  \caption{Real-LLM loop guard: redundant investigations executed (of $12$ duplicates),
  mean [min--max] over five seeded workloads per cell. The guard code is identical in every
  arm; only the injected per-lookup latency differs. ``Afforded'' = mean lookups fitting the
  $100\,$ms budget. The $110\,$ms counts are near-identical across models because no lookup
  fits the $100\,$ms budget at that latency, so the guard cannot fire; the residual redundancy
  is duplicates whose originals left the window, and the small model-to-model gap reflects each
  model's own deduplication.}
  \label{tab:llmguard}
  \begin{tabular}{lccc}
    \toprule
    Arm & Afforded & \texttt{gpt-5-nano} & \texttt{gpt-5-mini} \\
    \midrule
    (a) in-process guard   & $19$      & \textbf{0.0} [0--0] & \textbf{0.0} [0--0] \\
    (b) guard $+5\,$ms     & $19$      & 0.0 [0--0]          & 0.0 [0--0] \\
    (b) guard $+15\,$ms    & $17$      & 1.4 [0--4]          & 1.6 [0--4] \\
    (b) guard $+30\,$ms    & $8$--$9$  & 4.4 [2--7]          & 4.0 [2--6] \\
    (b) guard $+50\,$ms    & $4$       & 6.2 [5--8]          & 6.2 [5--8] \\
    (b) guard $+110\,$ms   & $0$       & 7.2 [6--9]          & 7.2 [6--9] \\
    (c) per-turn RAG       & ---       & 0.6 [0--3]          & 0.4 [0--2] \\
    \bottomrule
  \end{tabular}
\end{table}

\paragraph{Raising the budget does not buy the guard back.} Sweeping $B$ at the $110\,$ms
rung (\texttt{gpt-5-nano}): $B{=}150\,$ms affords $4$ lookups and leaks $6.2$; $250\,$ms
affords $8$ and leaks $4.4$; even $500\,$ms---half a second of memory dead time per
turn---affords ${\sim}15$ and still leaks $1.6$ [$0$--$4$]. The buyback is sublinear
because every affordable lookup costs a full round trip of turn time, while at in-process
speed the entire guard costs under a millisecond. The budget is not the constraint; the
round trip is.

\paragraph{The per-turn baseline is strong competition, and it clarifies the claim.}
Arm~(c), the industry-default pattern, performs well here---far better than the rationed
in-loop guard at cloud latency, and statistically indistinguishable from the in-process
guard on this workload's means (exact permutation $p{=}1.0$; the whole gap is one seeded
workload). Two readings follow. First, rationing is \emph{rational} at network latency: if
lookups cost $110\,$ms, falling back to one retrieval per turn beats a guard you cannot
afford, which is exactly why the industry converged on it. Second, per-turn coverage has a
structural leak: a turn-start memory dump cannot see duplicates that arise \emph{within}
the turn. Both models leak on the \emph{same} seeded workload (up to $3$ of $12$),
consistent with a structural rather than model-specific cause. The in-process guard is the
only arm at zero on every workload of both models, and its edge over per-turn widens as
agents take more steps per turn relative to their visible window. The claim these results
support is precise: latency does not change what the model knows; it changes what the
engineer can afford to wire around the model, and that alone moves task outcomes.

\paragraph{Cost.} The accuracy matrix consumed $0.82$M tokens across its eighty runs, plus
$0.29$M for the twenty gated-arm runs and $1.84$M for the sixty 25-turn-extension
runs; the
guard ladder and budget sweep $1.31$M across eighty-five
(\texttt{gpt-5-nano}/\texttt{-mini} only).

\section{The Embedding Bottleneck}
\label{sec:embedding}
Our measurements relocate the cost. With the store at ${\sim}100\,\mu\mathrm{s}$, the dominant
per-operation expense is \emph{embedding} (${\sim}200$--$400\,$ms via a network API). ``In-loop
memory is free'' is true of the store, not yet of the operation. The complete in-loop stack
therefore requires a \textbf{local embedder} alongside the in-process store. We measured this
stack directly: a 32M-parameter static embedder (\texttt{potion-retrieval-32M}) embeds at p50
$31.7\,\mu\mathrm{s}$, and with the in-process store the \emph{complete} embed-plus-store operation
lands at p50 $\mathbf{40\,\mu\mathrm{s}}$---roughly $5{,}000\times$ below the measured
network-embedding path ($202\,$ms)---placing the whole in-loop operation four orders of
magnitude under a reasoning step. (The frontier of \S\ref{sec:thesis} prices the same stack
at its conservative end, $c\approx116.6\,\mu\mathrm{s}$: it uses the $1{,}000$-item recall p50
of $84.9\,\mu\mathrm{s}$, the frontier's own store measurement alongside
Table~\ref{tab:live}'s clean-room $84.7\,\mu\mathrm{s}$, plus the local embed ($84.9 + 31.7 = 116.6$), where this benchmark's store op is measured on
a smaller working set; both numbers trace to the same artifacts.) We treat this as a finding: it identifies embedding
placement, not vector search, as the remaining obstacle to memory in the loop, and shows the
obstacle is already removable. The network embed is not only a time cost; it is a failure
window. In early accuracy-matrix runs, the $200$--$400\,$ms embed call held open a
concurrency window in our own tool wrapper in which same-turn writes collided on an id and
silently overwrote one another, destroying $47\%$ of saves until a one-line unique-id fix
(the pre-fix runs ship in the artifact). The fix is correct ids, not speed---but a
$32\,\mu\mathrm{s}$ local embed narrows the same window by four orders of magnitude, and the
incident is a measured example of what network-latency operations do to in-loop designs. (Reasoning-model wall-clock, roughly one to three minutes per
condition in \S\ref{sec:results}, is dominated by model latency, orthogonal to the store.)

\section{Related Work}
\label{sec:related}
\paragraph{Working memory and cognitive architectures.} Baddeley's model and its LLM mapping;
CoALA's working/long-term split and reasoning/retrieval/learning
actions~\citep{baddeley2000episodic,sumers2023coala}. These fix retrieval as an action but do
not ask how often it can run---the axis we add.

\paragraph{Agent memory systems.} MemGPT~\citep{packer2023memgpt} pages memory in/out of context
under memory pressure; Generative Agents~\citep{park2023generative} retrieve from a memory stream
scored by recency, importance, and relevance; Reflexion~\citep{shinn2023reflexion} and
Voyager~\citep{wang2023voyager} add episodic and procedural memory; surveys consolidate the
taxonomy~\citep{memorysurvey2026,humanmemai2025}. All optimize \emph{what} and \emph{whether} to
remember; none of these systems treat store latency as the variable gating the design space.
The emerging benchmark wave---MemBench~\citep{tan2025membench},
MemoryAgentBench~\citep{hu2025memoryagentbench}---evaluates what agents remember, at turn
granularity; what a store's speed lets the loop afford is not among its measurements.

\paragraph{Retrieval frequency and timing.} The RAG literature already names our variable one
level down: \citet{fan2024ragsurvey} catalogue \emph{retrieval frequency} as a serving-time
setting (one-time, every-$n$-tokens, every-token), and the Dynamics axis of the agent-memory
survey of \citet{hu2025memoryage} covers when retrieval is triggered. ProactAgent
\citep{cai2026proactive} names the contrast directly---``existing methods typically retrieve
memories passively, such as at task initialization or after each step''---and learns
\emph{when} to retrieve. All three treat frequency as a policy to choose or learn; none asks
what frequency an agent can \emph{afford}, which is where store latency enters and where our
frontier (\S\ref{sec:thesis}) sits.

\paragraph{Scratchpads and in-session working memory.} ReAct~\citep{yao2022react} keeps
working memory in-context; ``Empowering Working Memory''~\citep{empoweringwm2023} also
argues for externalizing it, into a cross-episode hub, but without a latency treatment. We
externalize it without leaving the loop, which only the latency argument makes viable.

\paragraph{Efficiency of in-loop retrieval.} \citet{searchagenteff2025} quantify in-loop latency
amplification and answer it at the serving layer: SearchAgent-X keeps retrieval interleaved and
unblocks it with priority-aware scheduling and non-stall retrieval.
AgentIR~\citep{yuan2026agentir} starts from the same observation---per-step memory queries
each ``a retrieval call that lives inside the agent's reasoning loop''---and answers with a cascade
retrieval substrate, treating latency as a serving constraint with no task-outcome claim;
AMV-L~\citep{bamidele2026amvl} manages tail latency through memory lifecycle management,
also serving-side. We answer at the substrate's \emph{location} instead
(\S\ref{sec:thesis}); the approaches compose. Industry ``memory-first''
guidance~\citep{mem0loop2026} does prescribe decoupling, with memory as a layer queried once per
turn; that is the prescription we reject.

\paragraph{Write-side hygiene.} Gating writes is itself established practice: SAGE gates
memory evolution on novelty \citep{wang2026sage}, governed-memory architectures ration
writes in production multi-agent stacks \citep{taheri2026governed}, retrieval
restructuring beyond flat RAG reorganizes what agents read back \citep{hu2026beyondrag},
and Entity-Collision stratifies how retrieval lift is attributed under near-duplicate
entities \citep{deng2026entitycollision}. None of these measures what a write gate does
to task outcome under a fixed, disclosed read policy against an instructed restatement
baseline; that measurement is \S\ref{sec:results}'s.

\paragraph{Latency and outcomes.} \citet{kang2025winfast} show inference latency flipping
task outcomes when an external clock punishes slowness (real-time games, trading)---latency
as a race against the world. Our mechanism needs no clock: store latency degrades the
agent's \emph{information state}, so it repeats work even with unlimited time. Closest on the
systems side, \citet{omri2026agentmem} characterize agent-memory workloads and relate
memory-serving latency to task accuracy across many systems; but every system they measure
sits at ${\sim}0.1\,$s or above, and they vary neither retrieval frequency nor latency by
intervention, so the relation they report holds \emph{across} designs rather than the
within-design causal move we isolate. Practitioner
writing has reached the intuition---a vendor post benchmarks sub-millisecond retrieval
across reasoning steps~\citep{ryjox2026speed} (its own number measured, competitor latencies
simulated, wall-clock only), and
Redis's agent-memory guidance recommends in-memory storage where ``response time compounds
across multiple reasoning steps''~\citep{redis2026memory}, stopping at the network
hop---without a frequency axis, a parity criterion, or an outcome experiment.

\paragraph{Extended cognition.} \citet{clark1998extended} and recent AI
bridges~\citep{aithought2025,riva2025invisible} invoke extended cognition descriptively.
Mainen's Library Theorem~\citep{mainen2026library} is the closest formal predecessor: parity
for file-backed agents with \emph{organization} as the gate (\S\ref{sec:parity} engages its
argument against the working-memory framing); \citet{perrier2025operationalising}
operationalises extended cognition for corporate knowledge, with access efficiency inside
its knowledge metric but no agents and no latency threshold; a unified
review~\citep{zhou2026externalization} maps externalization of memory, skills, and
protocols. We use the parity principle prescriptively, as a latency criterion
(\S\ref{sec:parity}).

\paragraph{What remains unclaimed.} Each ingredient of our claim now exists separately:
AgentIR's in-loop cost arithmetic, Kang et al.'s latency-flips-outcomes under a clock, the
Library Theorem's parity for external stores, industry's sub-millisecond intuition. To our
knowledge, no prior work holds the conjunction---store latency as the gate on retrieval
frequency, parity as its criterion---and none demonstrates the causal chain \emph{slower
store $\rightarrow$ fewer affordable checks $\rightarrow$ redundant actions} with only the
store's answer speed moved. That conjunction is this paper.

\section{Discussion}
\label{sec:discussion}
\paragraph{A decision rule for practitioners.} The cost model of \S\ref{sec:thesis} answers
the affordability half: per-step access is viable when $f\cdot c \ll r$, and with an
in-process store and local embedder ($c\approx116.6\,\mu\mathrm{s}$) that holds at any realistic
frequency. The results of \S\ref{sec:causalresults} answer the task half. Per-turn retrieval
remains adequate when each turn needs only facts settled in earlier turns---arm~(c) performed
respectably for exactly this reason. In-loop access becomes necessary when correctness
depends on state that changes \emph{within} the turn: duplicate work, contradictions,
constraints discovered mid-reasoning. The rule of thumb: if a mistake can arise between two
retrievals, the retrieval interval is too long---and only a store priced in microseconds lets
that interval shrink to a single step.

\paragraph{Allocate, don't connect.} The latency inversion reframes what an agent's memory
\emph{is}. A networked store is infrastructure: provisioned, shared, connected to, kept
alive. An in-process store is a data structure: allocated per agent, per task, even per turn,
and discarded when done---the clean-room benchmark's ${\sim}60{,}000$ create-fill-destroy
lifecycles per second is the point, not a stunt. Agent frameworks today expose memory as a
service client configured at startup; these results argue for exposing it as a loop primitive
in the same category as the scratchpad, so that guard patterns like \S\ref{sec:causalresults}'s
are library one-liners rather than architecture decisions.

\paragraph{Division of labor with the model.} The sharpest lesson of
\S\ref{sec:causalresults} is what the store did \emph{not} do: it did not make the model
smarter. The models still failed to recognize seven of twelve disguised duplicates on their
own. The system stopped executing them because a code guard could afford to check a reliable
store at every step. Reliability was engineered into the scaffold around the model rather
than prompted into the model---a design stance that only becomes available when checking
costs microseconds.

\paragraph{What else free per-step access unlocks.} The loop guard is deliberately the
simplest member of a family (\S\ref{sec:thesis}): novelty checks on every observation,
grounding of claims as they are produced, working sets that outlive the window, and shared
in-process blackboards for co-located agents. Each is one \texttt{add} or \texttt{match} per
step, affordable only above the feasibility frontier. We demonstrate the guard; the rest are
measured next steps, not speculation about capability.

\section{Limitations}
\label{sec:limitations}
\begin{figure}[t]
  \centering
  \includegraphics[width=\linewidth]{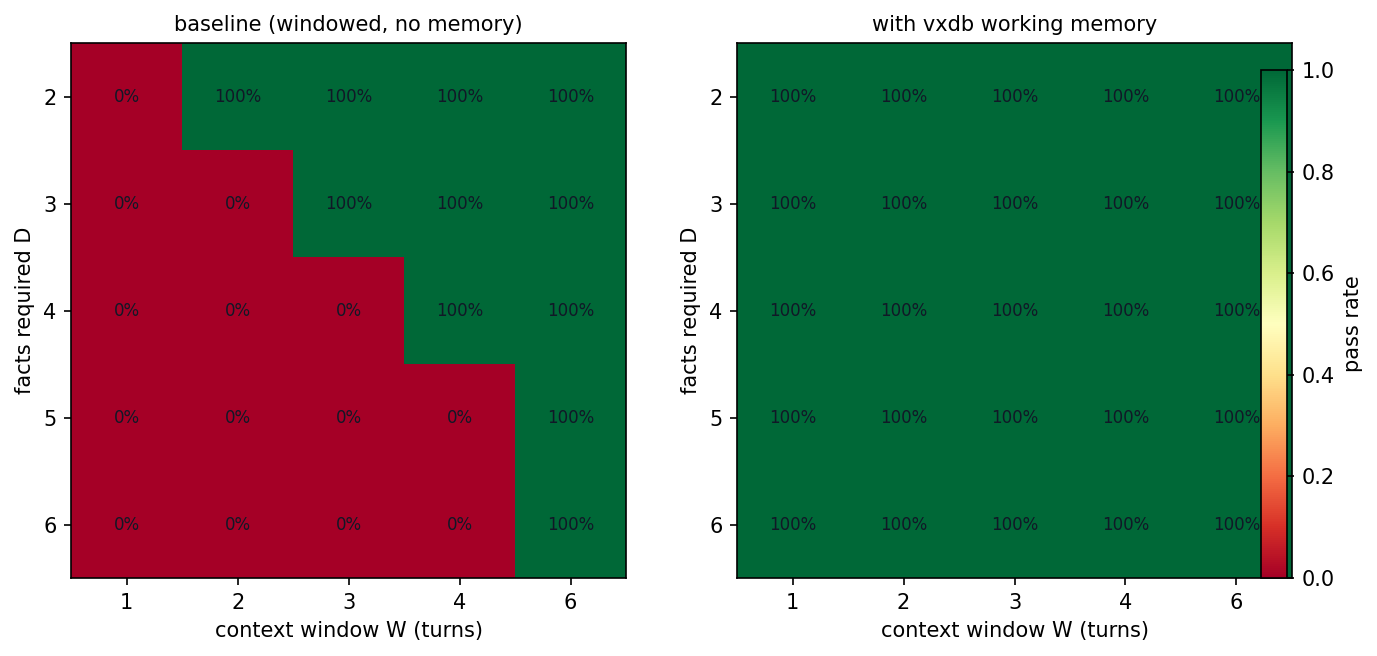}
  \caption{The window cliff (sensitivity analysis): task pass rate across fact span
  $D$ (one fact per turn) $\times$ context window $W\in\{1,2,3,4,6\}$ turns, averaged over three task families (deterministic
  mechanism check). The windowed baseline (left) fails exactly where $D$ exceeds $W$; the
  same agent with working memory (right) passes every configuration.}
  \label{fig:sweep}
\end{figure}

Two task families (constraint recall, loop-guard dedup) and one window size in the main
results; a deterministic window$\times$facts sweep across three task families (no model in
the loop) locates the failure boundary at fact span $D >$ window and serves as sensitivity
analysis (Figure~\ref{fig:sweep}). The
trip task is small---five facts, six turns---and the restatement baseline wins at that
scale (the gated memory
arm ties it on two of four models); the 25-turn extension measures the rent on the turns
axis (the cumulative gap widens monotonically and the store undercuts restatement on every
model by turn 25, Figure~\ref{fig:tokencurve}), while its breakdown at larger working
sets (the facts axis) follows from the rent argument
(\S\ref{sec:background}) and remains argued, not measured here.
The gate bounds duplication, not store size: \texttt{gpt-5}'s gated stores in the 25-turn
runs still grew to $9$--$12$ items on incidental saves, past the fixed $k{=}8$ read, so
keeping a long-horizon store inside a fixed read window remains open.
The gated arm ran after the other Table~\ref{tab:matrix} cells; a later single-sitting rerun of all five
conditions ($100$ fresh runs, shipped in the artifact) reproduced the gated cells exactly
on three of four models ($5.0$, $4.8$, $5.0$). The fourth, \texttt{gpt-5-nano}, dropped on
runs that saved facts but never called recall, the \S\ref{sec:method} read-policy class no
write-side check addresses; ungated means moved between sittings too (keyword grader: $3.4$--$4.2$ vs
$3.6$--$4.8$).
Constraint grading is keyword-based and deliberately loose, mitigated by a second,
rubric-based LLM grade---same provider family, so agreement is corroboration, not
independence; a cross-family grader is future work. The loop guard matches the model's
extracted action line rather than raw alert
prose; the offline calibration (\S\ref{sec:causalmethod}) shows why---a static embedder
ranks same-template non-duplicates above true rewordings---and richer matchers over raw
prose remain untested here. The budget policy is deliberately naive (guards the earliest
affordable steps; no caching, no selective guarding), so the reported leak rates are a floor
on what smarter rationing could recover. The networked baseline is measured on a
\emph{loopback} Qdrant (\S\ref{sec:results}), so a WAN/multi-region deployment and an
end-to-end agent run against a live cloud store would strengthen it further; the delayed
arms inject measured round trips rather than calling one. Accuracy runs are repeats, not
seeded replications (the harness exposes no seed); LLM tool-use variance is real and
reported as means with ranges, and the memory condition's misses trace to its read policy
(one recall per run through a fixed $k{=}8$ window), never to store failures; $k$ is a
design constant here, not swept. The gated arm measures how much the write side alone
recovers (most of the gap, \S\ref{sec:results}) while the rest of the read-policy axis
stays untested: the ask-only trigger
of \S\ref{sec:method}, the width and repetition of reads, and above all who fires the
recall (the model's tool-calling disposition or the loop's wiring) all remain open. The
ungated memory scores should be read as the floor under the weakest of those policies, and
the gated arm as evidence that the floor is not the mechanism's ceiling. The
constitutive-cognition claim is functional, not phenomenal
(\S\ref{sec:parity}).

\section{Conclusion}
\label{sec:conclusion}
Memory has stayed outside the agent loop for one reason, and it is not architecture. At
network latency, per-step access costs seconds of dead time per turn, so the field engineered
around the store: scheduling its calls, rationing them to turn boundaries. We showed the
premise is contingent---an in-process store answers three orders of magnitude faster---and,
more importantly, that it is causal: holding a per-turn memory budget fixed and moving only
the store's answer speed takes a scripted agent from zero redundant actions to ten of ten,
and traces a monotone dose-response in real LLM agents, from $0.0$ redundant investigations
at in-process speed to $7.2$ of $12$ at cloud latency. Read through the parity principle as a latency
budget, a store that fast becomes \emph{eligible} to be working memory the agent has rather
than a database it consults---and a loop wired to consult it on every step is what realizes
the eligibility. The last obstacle, network embedding, falls to a measured $40\,\mu\mathrm{s}$
complete operation with a small local embedder; the full stack exists today. Retrieval
frequency---per-turn versus per-step---is now a design variable rather than a constraint, and
the interesting question is no longer whether an agent can afford to check its memory at
every step, but what agents will be built by engineers who know that it can.

Three studies follow directly. A \emph{long-horizon study}: the guard's edge over per-turn
retrieval grows with steps per turn and duplicates within a turn, and measuring that growth
curve across task families and model families is the natural next experiment. A
\emph{confusion-resistance study}: everything here uses ephemeral, task-scoped stores; how
in-loop access interacts with long-lived, growing memories---where stale and near-duplicate
entries compete---is open. And an \emph{artifact study}: the guard patterns of
\S\ref{sec:causalresults} as library one-liners, so that per-step memory is an allocation
decision rather than an architecture decision.

This paper settles only the temporal half of the memory question: \emph{when} memory
participates in reasoning. It is a first step toward treating memory as a cognitive resource
for language agents rather than auxiliary infrastructure. Two questions follow it. One is the
lifecycle of what an agent keeps: how it prioritizes, forgets, and decides what to trust. The
other is the organization of what it keeps: structured, symbolic representations suited to
individual and collective cognition. Pursued together, they point toward memory as a
fundamental computational substrate for intelligent agents.

\ifanon\else
\section*{Acknowledgments and Disclosure of Funding}
Yusuf Khan is the developer of vxdb, the in-process vector store used as the
measurement instrument in this work. This work received no external funding.
\fi

\bibliographystyle{plainnat}
\bibliography{references}

\end{document}